\documentclass[11pt]{article}
\usepackage{coling2020}
\usepackage{times}
\usepackage{url}
\usepackage{latexsym}

\usepackage{algorithm, algorithmic}
\usepackage{multirow}
\usepackage{multicol}
\usepackage{adjustbox}
\usepackage{diagbox}
\usepackage{bm}
\usepackage{graphicx}
\usepackage{amsfonts}		% Additional math fonts
\usepackage{xspace}
\usepackage{listings}
\usepackage{verbatim}
\usepackage{varwidth}
\usepackage{tabularx}
\usepackage{subfig}
\usepackage{array}
\usepackage{makecell}

\usepackage{microtype}
\usepackage{amsmath}
\usepackage{xcolor}

\definecolor{darkspringgreen}{rgb}{0.05, 0.5, 0.06}
\newcommand{\orw}[1]{\textcolor{green}{\textbf{#1}}}
\newcommand{\mow}[1]{\textcolor{red}{\textbf{#1}}}

\DeclareMathOperator*{\argmax}{argmax} % thin space, limits underneath in displays

\colingfinalcopy % Uncomment this line for the final submission

\title{A Geometry-Inspired Attack for Generating Natural Language Adversarial Examples}

\author{Zhao Meng, Roger Wattenhofer \\
Department of Electrical Engineering and Information Technology\\
  ETH Zurich, Switzerland \\
  \texttt{\{zhmeng, wattenhofer\}@ethz.ch} \\}

\date{}

\begin{document}
\maketitle
\begin{abstract}
Generating adversarial examples for natural language is hard, as natural language consists of discrete symbols, and examples are often of variable lengths. In this paper, we propose a geometry-inspired attack for generating natural language adversarial examples. Our attack generates adversarial examples by iteratively approximating the decision boundary of Deep Neural Networks (DNNs). Experiments on two datasets with two different models show that our attack fools natural language models with high success rates, while only replacing a few words. Human evaluation shows that adversarial examples generated by our attack are hard for humans to recognize. Further experiments show that adversarial training can improve model robustness against our attack.
\end{abstract}

\section{Introduction}

Although Deep Neural Networks (DNNs) have been successful in many machine learning applications~\cite{sentence_classification,machine_comprehension,resnet}, 
researchers have demonstrated that DNNs are remarkably vulnerable to adversarial attacks, which generate adversarial examples by adding small perturbations to the original input 
\cite{adv_original3,adv_original,adv_original2}. 
Adversarial examples are essential as they showcase the limitations of DNN models. Like humans, good DNN models should be robust to small perturbations to inputs. If a DNN model judges two almost identical inputs differently, one must profoundly question the quality of the DNN. As such adversarial examples are more than just a gimmick: they are a proof of the fundamental limitations of a DNN model.

Previous research on adversarial attacks has been largely focused on images, e.g., ~\cite{threat}. 
In this paper, we study how to adversarially attack natural language models. 
Generating adversarial examples for natural language is fundamentally different from generating adversarial examples for images. Images live in a continuous universe, where one can simply change pixel values. 
Natural language sentences and words on the other hand are typically discrete. 
This discrete nature makes it difficult to apply existing attacks from the image domain directly to natural language, as an arbitrary point in the input space is unlikely to correspond to a valid natural language sentence or word. Moreover, inputs of natural language to DNNs are of variable lengths, which further complicates generating adversarial examples for natural language.

Despite these obstacles, researchers have proposed various attacks to generate adversarial examples for natural language. \newcite{machine_reading_adv} manage to fool a DNN model for machine reading by adding sentences to the original texts. \newcite{natural_adv} generate adversarial examples for natural language by using an autoencoder. \newcite{hotflip} propose a gradient-based attack to generate adversarial examples in the granularity of individual characters. \newcite{adv_mha} generate fluent adversarial examples using Metropolis-Hastings sampling. \newcite{pwws} combine several heuristics to generate adversarial examples.

However, all these methods do not address the ``geometry'' of DNNs, which has been shown to be a useful approach in the image domain~\cite{deepfool,cure,sparsefool}.
In this paper, we propose a geometry-inspired attack\footnote{Code: \url{https://github.com/zhaopku/nlp_geometry_attack}} for generating natural language adversarial examples. Our attack generates adversarial examples by iteratively approximating the decision boundary of DNNs. 
We conduct experiments with Convolutional Neural Networks (CNNs) and Recurrent Neural Networks (RNNs) on two text classification tasks: the IMDB movie review dataset, and AG's News dataset.
Experimental results show that our attack fools the models with high success rates while keeping the word replacement rates low. We also conduct a human evaluation, showing that adversarial examples generated by our attack are hard for humans to recognize. Further experiments show that model robustness against our attack can be achieved by adversarial training. 
\section{Related Work}

Despite the success of Deep Neural Networks (DNNs) in many machine learning applications~\cite{sentence_classification,resnet}, researchers have revealed that such models are vulnerable to adversarial attacks, which fool DNN models by adding small perturbations to the original input~\cite{adv_original}. The vulnerability of DNN models poses threats to many applications requiring high-level security. For example, in the image domain, a small error in a self-driving car could lead to life threatening disaster. For natural language, a machine might misunderstand a meaning, coming to a wrong conclusion. Researchers have also shown that a universal trigger could lead a system to generate highly offensive language~\cite{universal}.

Previously, researchers have developed various adversarial attacks for fooling DNN models for images. \newcite{adv_original} propose Fast Gradient Signed Method (FGSM), which aims to maximize the loss of the model with respect to the correct label. Projected Gradient Descent (PGD)~\cite{pgd} can be viewed as a multi-step version of FGSM. In each step, PGD generates a perturbation using FGSM, and then projects the perturbed input to an $l_\infty$ ball. While these gradient-based methods are effective, researchers also show that leveraging geometry information of DNNs can be helpful. \newcite{deepfool} and \newcite{sparsefool} generate adversarial examples by iteratively approximating the decision boundary of DNNs. 

Although many methods have been proposed for generating adversarial examples for images, little attention has been paid to generating adversarial examples for natural language. Generating adversarial examples for natural language is fundamentally different from generating adversarial examples for images. On the one hand, while pixel values of images are continuous, natural language consists of sequences of discrete symbols. Moreover, natural language sentences and words are often of variable lengths. Hence, existing adversarial attacks designed for images cannot be directly applied to natural language.

Despite obstacles, researchers have proposed various methods for generating adversarial examples for natural language.
Based on the granularity of adversarial perturbations, adversarial attacks for natural language models can be divided into three categories: character level, word level and sentence level.

\subsection{Character Level}

Character-level adversarial attacks for natural language models generate adversarial examples by modifying individual characters of the original example. \newcite{hotflip} propose HotFlip, which uses gradient information to swap, insert, or delete a character in an original example. \newcite{bugger} generate adversarial examples by first selecting important words, and then modifying characters of the selected words.

Although character-level adversarial attacks for natural language are effective, such methods suffer from the problem of perceptibility. Humans are likely to recognize adversarial examples generated by these methods, as changing individual characters of texts often results in invalid words. Furthermore, character-level adversarial attacks are easy to be defended against. Using a simple spell checking tool to preprocess inputs can defend a DNN model against such attacks.

\subsection{Word Level}
Word-level adversarial attacks generate adversarial examples for natural language by changing words of the original example. \newcite{genetic} propose a genetic attack, in which they replace original words with their synonyms by iteratively applying a genetic algorithm. \newcite{adv_mha} generate adversarial examples for natural language by leveraging Metropolis-Hastings Sampling. \newcite{pwws} leverage several heuristics to generate word-level adversarial examples. \newcite{universal} propose a universal attack, in which a fixed, input-agnostic sequence of words triggering the model to make a specific prediction is prepended to any example from the dataset. They search such universal triggers by leveraging gradient information.

\subsection{Sentence Level}
While most researchers focus on character/word-level attacks, some researchers propose to fool DNN models for natural language with sentence-level attacks. \newcite{machine_reading_adv} propose to fool a machine reading model by adding an additional sentence to the original texts. However, their method requires heavy human engineering. \newcite{rewrite} generate adversarial examples by rewriting the entire sentence with an encoder-decoder model for syntactically controlled paraphrase generation.

\bigskip
All these methods, however, do not address the geometry of DNNs although such information has been proven useful for generating adversarial examples for images.
In this paper, we propose a geometry-inspired word-level adversarial attack for generating natural language adversarial examples. The rest of this paper is organized as follows. Section~\ref{sec:method} describes our attack. Section~\ref{sec:exp} details the experimental settings as well as results. Section~\ref{sec:conclusion} gives conclusions and insights for future work.

\section{Methodology}\label{sec:method}

Our attack is a white-box attack in that the attacker has access to the architecture and parameters of the victim model. The attack crafts natural language adversarial examples by replacing original words with their synonyms. Specifically, our attack can be divided into two steps: word selection and synonym replacement. In each iteration, the attack first selects a word from the original text, and then replaces the selected word with one of its synonyms to craft an adversarial example. The remainder of this section gives the details of our attack.

\subsection{Word Selection Strategy}

A crucial step in generating text adversarial examples is to find which word of the original example to replace. We follow previous work by ranking words with their saliency scores~\cite{saliency_0,saliency_1,pwws}. The saliency score of word $w_j$ is obtained by computing the decrease of true class probability after replacing $w_j$ with an out-of-vocabulary word ${u}$, embeddings of which are initialized to all zeros during training.

Specifically, we have

\begin{align}
    \bm X &= w_0, w_1, \dots, w_j, \dots, w_{N-1} \\
    \bm X^\prime &= w_0, w_1, \dots, u, \dots, w_{N-1}
\end{align}

\noindent where $\bm X^\prime$ is obtained by replacing word $w_j$ of the original example $\bm X$ with out-of-vocabulary word ${u}$. Let $y$ be the ground truth label of original example $\bm X$. The saliency score $S_j$ for word $w_j$ is given by

\begin{align}
    S_j = P(y|\bm X) - P(y|\bm X^\prime)
\end{align}

A higher saliency score indicates the corresponding word is of more importance for predicting the true class. Hence, the word with the highest saliency score in candidate set $\mathbb{C}$ will be selected for replacement. We build the candidate set $\mathbb{C}$ from words of the original example $\bm X$ and then exclude all out-of-vocabulary words and punctuations.
\begin{algorithm}[!t]

\caption{Adversarial Attack}
\begin{algorithmic}[1]
\STATE \textbf{input:} Example $\bm X = w_0, w_1, \dots, w_{N-1}$, true label $y$, classifier $f$ with text encoder $\texttt{Encoder}$ and feed forward layer $\texttt{FFNN}$.
\STATE \textbf{output:} Adversarial example $\hat{\bm X}$.
% \State $\r\gets \mathbf{0}$
\STATE Initialize $\bm X_0\gets \bm X$, candidate set $\mathbb{C} = \{w_0, w_1, \dots, w_{K-1}\}$, $i \gets 0$, projections $\mathbb{P} \gets \{\}$, $i \gets 0$.
\WHILE{$\mathbb{C} \neq \emptyset$}
\FOR{$w_k \in \mathbb{C}$}
\STATE $S_k$ $\gets$ WordSaliency($\bm X_i$, $w_k$) // compute word saliency
\ENDFOR
\STATE $k^* \gets \argmax_k S_k,$  where $w_k \in \mathbb{C} $
\STATE $\mathbb{Q}_k^* \gets \{w_{k^*}^0, w_{k^*}^1, \dots, w_{k^*}^{M_j-1}\}$ // synonym set of $w_{k^*}$
\STATE $\bm v_i \gets \texttt{Encoder}(\bm X_i)$
\STATE $\bm b_i \gets \texttt{DeepFool}(\bm v_i, \texttt{FFNN})$
\STATE $\bm r_i \gets \bm b_i - \bm v_i$
\STATE $\bm u_i \gets \frac{\bm r_i}{||\bm r_i||}$
\FOR{$m = 0$ to $M_j-1$}
\STATE Craft $\bm X_i^m$ by replacing $w_{k^*}$ with $w_{k^*}^m$
\STATE $\bm v_i^m \gets \texttt{Encoder}(\bm X_i^m)$
\STATE $\bm d_i^m \gets \bm v_i^m - \bm v_i$
\STATE $\bm p_i^m \gets \texttt{Projection}(\bm d_i^m, \bm r_i)$
\STATE $\mathbb{P} \gets \mathbb{P} \cup \{\bm p_i^m\}$
\ENDFOR
\STATE $m^* \gets \argmax_m  (\bm p_i^m \cdot \bm u_i)$, where $\bm p_i^m \in \mathbb{P}$
\STATE $\bm X_{i+1} \gets \bm X_i^{m^*}$
\IF{$f(\bm X_{i+1}) \neq f(\bm X)$}
\STATE $\hat{\bm X} \gets \bm X_{i+1}$
\STATE break
\ENDIF
\STATE $\mathbb{C} \gets \mathbb{C} - \{w_{k^*}\}$
\STATE $i \gets i+1$
\ENDWHILE
\STATE \textbf{return} $\hat{\bm X}$
\end{algorithmic}

\label{alg:attack}
\end{algorithm}

\begin{figure}
    \centering
    \includegraphics[width=0.3\columnwidth]{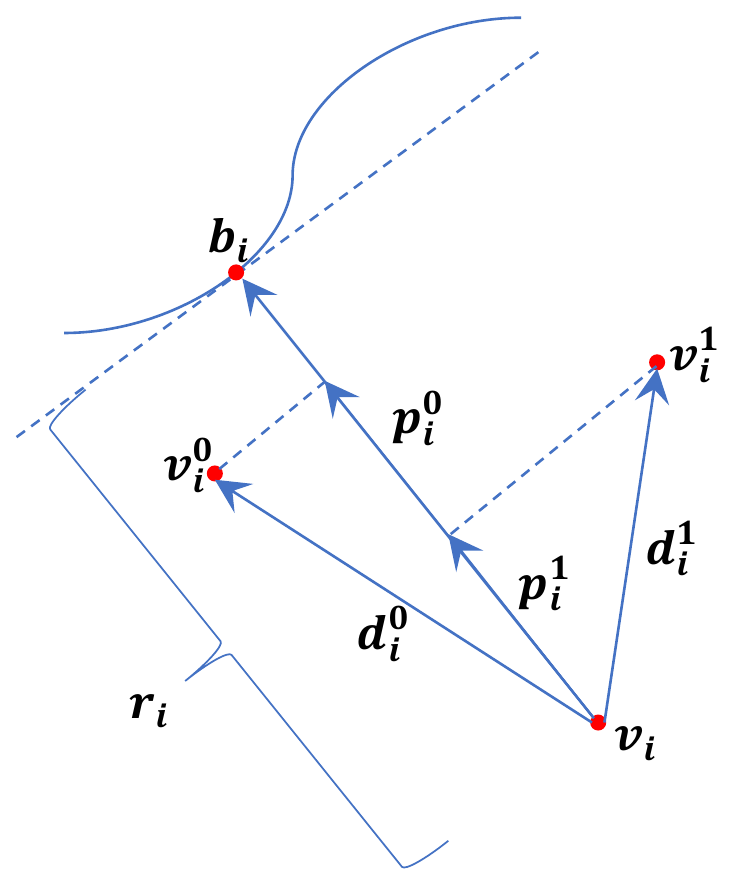}
    \caption{Illustration of iteration $i$ in our attack: $\bm v_i$ is the original text vector. The curved line on top is the decision boundary, with $\bm b_i$ being the closest point on the decision boundary to $\bm v_i$. %The synonym corresponding to the largest projection is chosen.    
    $\bm v_i^0$ and $\bm v_i^1$ are text vectors obtained by replacing word $w_{k^*}$ with its synonyms $w_{k^*}^0$ and $w_{k^*}^1$, respectively. $\bm p_i^0$ ($\bm p_i^1$) is the projection of $\bm d_i^0$ ($\bm d_i^1$) onto $\bm r_i$. In this iteration, $w_{k^*}^0$ is chosen over $w_{k^*}^1$ as $||\bm p_i^0|| > ||\bm p_i^1||$. We also have $z_i^{max} = ||\bm p_i^0||$ in this example.}
    \label{fig:illustration}
\end{figure}

\subsection{Synonym Replacement Strategy}\label{sec:synonym}

Before going into the details of our synonym replacement strategy, we first clarify our assumptions on model architectures. For text classification tasks, a model can be divided into a text encoder $\texttt{Encoder}$ and a feed forward layer \texttt{FFNN}. Specifically, a text encoder encodes an input $\bm X$ into a fixed-size vector $\bm v$. Choices of such encoders include RNNs, CNNs~\cite{sentence_classification}, etc. A feed forward layer then takes the fixed-size vector $\bm v$ as input for classification. A fully connected network followed by a softmax activation layer is common for feed forward layers.

Our attack iterates over the candidate set $\mathbb{C}$ to generate adversarial examples. In each iteration, we first compute word saliency score $S_k$ for each candidate word $w_k \in \mathbb{C}$. We derive the synonym set $\mathbb{Q}_{k^*} = \{w_{k^*}^0, w_{k^*}^1, \dots, w_{k^*}^{M_j-1}\}$ using \texttt{WordNet}\footnote{\url{https://wordnet.princeton.edu/}} for candidate word $w_{k^*}$, which has the largest saliency score $S_{k^*}$ in the current iteration.

We then use geometric information to select the best synonym of $w_{k^*}$ for replacement. 
Given a DNN classifier consisting of text encoder \texttt{Encoder} and feed forward layer \texttt{FFNN}, we first use \texttt{Encoder} to compute the text vector $\bm v_i$ of $\bm X_i$, which is the example before replacement at iteration $i$.
We then find the nearest point $\bm b_i$ on the decision boundary of \texttt{FFNN} by leveraging the \texttt{DeepFool} algorithm~\cite{deepfool}. Next, we compute $\bm r_i$, which originates from text vector $\bm v_i$ to decision boundary point $\bm b_i$. 

For each synonym $w_{k^*}^m \in \mathbb{Q}_{k^*}$, example $\bm X_i^m$ is obtained by replacing $w_{k^*}$ with $w_{k^*}^m$. We compute text vector $\bm v_i^m$ by feeding $\bm X_i^m$ into the \texttt{Encoder}. We obtain $\bm p_i^m$ by projecting $\bm d_i^m$, which is the vector originating from $\bm v_i$ to $\bm v_i^m$, onto $\bm r_i$. 
A new example $\bm X_{i+1}$ is crafted by replacing original word $w_{k*}$ with its synonym $w_{k^*}^{m*}$, which corresponds to the largest projection $z_i^{max}$, where $z_i^{max} = \bm p_i^{m^*} \cdot \bm u_i$, with $\bm u_i$ being the unit direction vector of $\bm r_i$.
Our intuition is that a text vector with larger projection on $\bm r_i$ is closer to the decision boundary. We assign $\bm X_i$ to $\bm X_{i+1}$ directly and continue to the next iteration if $z_i^{max}$ is negative (which indicates $\bm p_i^{m^*}$ is in the opposite direction of $\bm u_i$ and $\bm r_i$). Figure~\ref{fig:illustration} illustrates our synonym replacement strategy. The algorithm stops under the condition that the model is fooled or the candidate set $\mathbb{C}$ is exhausted. We give details of our attack in Algorithm~\ref{alg:attack}.
\begin{table*}
\LARGE
\resizebox{1.0\textwidth}{!}{%
\begin{tabular}{|c|c|c|c|c|}\hline
\textbf{Example} & \textbf{Predictions} & \textbf{Replacements} & \textbf{Distance} & \textbf{True Class Prob}\\\hline \hline

\multicolumn{1}{|m{0.7\textwidth}|}{Obviously, most of the budget was put into the dinosaurs, and although there is a fair share of 
them, there's not nearly enough to \orw{save} (\mow{preserve}) us from our \orw{boredom} (\mow{ennui}). These human characters are only there to scream, run around, and mutter these poorly-written and verbose speeches about survival. And \orw{unfortunately} (\mow{regrettably}), not nearly enough of them get eaten by the dinosaurs. Overall , ``planet of the dinosaurs'' is not a film I plan on seeing again.} & \texttt{Negative} $\rightarrow$ \texttt{Positive} &  \makecell{\orw{boredom} $\rightarrow$ \mow{ennui} \\ \orw{save} $\rightarrow$ \mow{preserve}  \\  \orw{unfortunately} $\rightarrow$ \mow{regrettably}} & \makecell{0.80 $\rightarrow$ 0.56 \\ 0.56 $\rightarrow$ 0.25 \\  0.25 $\rightarrow$ -0.14} & \makecell{84.19\% $\rightarrow$ 76.55\% \\ 76.55\% $\rightarrow$ 62.62\% \\ 62.62\% $\rightarrow$ 42.93\%} \\\hline

\multicolumn{1}{|m{0.7\textwidth}|}{Screening as part of a series of funny shorts at the sydney gay and lesbian mardi gras film festival, this film was \orw{definitely} (\mow{unquestionably}) a highlight. The script is \orw{great} (\mow{smashing}) and the direction and acting was terrific. As another posting said, the actors' comedic timing really made this film. Lots of fun.} & \texttt{Positive} $\rightarrow$ \texttt{Negative} & \makecell{\orw{great} $\rightarrow$ \mow{smashing} \\ \orw{definitely} $\rightarrow$ \mow{unquestionably}}  & \makecell{1.56 $\rightarrow$ 0.01 \\ 0.01 $\rightarrow$ -0.79} & \makecell{96.37\% $\rightarrow$ 50.12\% \\ 50.12\% $\rightarrow$ 15.88\%}  \\\hline

\multicolumn{1}{|m{0.7\textwidth}|}{This is the first movie I have watched in ages where I actually ended up fast forwarding through the \orw{tedious} (\mow{wordy}) bits which there are plenty of. Very ordinary movie. I'm glad I missed it at the movies \& got a 2 for 1 video deal which included this movie instead.} & \texttt{Negative} $\rightarrow$ \texttt{Positive} &  \orw{tedious} $\rightarrow$ \mow{wordy} & 2.45 $\rightarrow$ -1.89 & 99.42\% $\rightarrow$ 40.21\% \\\hline \hline

\multicolumn{1}{|m{0.7\textwidth}|}{\orw{Ready} (\mow{Prepare}) to \orw{bet} (\mow{depend}) on alternative energy? Well, think again when oil prices rise, public interest in alternative energy often does, too. But the logic is evidently escaping wall street. } & \texttt{Business} $\rightarrow$ \texttt{Sci/Tech} & \makecell{\orw{bet} $\rightarrow$ \mow{depend} \\ \orw{ready} $\rightarrow$ \mow{prepare}} & \makecell{2.65 $\rightarrow$ 0.34 \\ 0.34 $\rightarrow$ -2.19} & \makecell{99.65\% $\rightarrow$ 68.03\% \\ 68.03\% $\rightarrow$ 0.51\%} \\\hline

\multicolumn{1}{|m{0.7\textwidth}|}{Convicted spammer gets nine years in \orw{slammer} (\mow{jailhouse}) A brother and sister have been convicted of three felony charges of sending thousands of junk e-mails; 
one of them was sentenced to nine years in prison, the other was fined \$ 7,500. } & \texttt{Sci/Tech} $\rightarrow$ \texttt{Business} &  \orw{slammer} $\rightarrow$ \mow{jailhouse} & 2.64 $\rightarrow$ -0.98 & 99.77\% $\rightarrow$ 8.67\% \\\hline

\multicolumn{1}{|m{0.7\textwidth}|}{Osaka school \orw{killer} (\mow{slayer}) of 8, Yakuza boss executed Yokyo - Mamoru Takuma, convicted for murdering eight children at an Osaka elementary school in 
2001, has been executed, informed sources said Tuesday.} & \texttt{World} $\rightarrow$ \texttt{Sports} &  \orw{killer} $\rightarrow$ \mow{slayer} & 1.59 $\rightarrow$ -1.60 & 98.20\% $\rightarrow$ 1.60\% \\\hline

\end{tabular}
}
\caption{Adversarial examples from our attack. Irrelevant parts of an example are omitted for simplicity. The first three examples are from the IMDB dataset, and the last three are from AG's News dataset. We use LSTM-based RNN for both datasets. \orw{Green} words are original words, while \mow{red} words are replaced words. \textbf{Predictions}: model predictions before and after the attack. \textbf{Replacements}: word replacements. \textbf{Distance}: changes of distance from text vector to decision boundary. \textbf{True Class Prob}: changes of true class probability as original words being replaced. ``{True Class}" refers to the true class of the original example.
}
\label{tab:example}

\end{table*}
\begin{table}[!h]
\centering
\resizebox{0.6\linewidth}{!}{
\begin{tabular}{|l||c|c|c|c|c|}
\hline
\textbf{Dataset}                    & \textbf{\#Train}        & \textbf{\#Test}   & \textbf{\#Classes} & \textbf{Avg. \#Words} & \textbf{Max. \#Words}     \\ \hline \hline
{IMDB}                       & 25,000                  & 25,000    & 2      &   258    & 600 \\ \hline
{AG's News}                  & 120,000                 & 7,600     & 4      &   43     & 248 \\ \hline
\end{tabular}
}
\caption{Statistics of datasets. Note that we limit the maximum number of words per example for the IMDB dataset to 600, while we do not limit the maximum number of words for the AG’s News dataset.}
\label{tab:dataset}
\end{table}

\section{Experimental Results}\label{sec:exp}

We elaborate our experiments in this section. Section~\ref{sec:exp-setup} details the experimental settings. Section~\ref{sec:exp-attack} describes the results of adversarial attacks. We conduct a human evaluation in Section~\ref{sec:exp-human} to understand the perceptibility of our adversarial perturbations. Section~\ref{sec:exp-training} gives the results of adversarial training, which we found can improve the robustness of DNN models against our attack. 

\subsection{Setup}\label{sec:exp-setup}

We describe our experimental setup, including datasets and models in this subsection. We test our attack on two datasets with two different models. 

% \subsubsection{Datasets}

\noindent\textbf{Datasets} We conduct our experiments on two English datasets for text classification. Specifically, we have
\begin{itemize}
    \item \textbf{IMDB}\footnote{\url{https://ai.stanford.edu/~amaas/data/sentiment/}}~\cite{imdb_dataset}: The IMDB dataset is a large dataset for binary sentiment classification. Each example in the dataset is a movie review. The classification label is \texttt{positive}/\texttt{negative}. Both labels are equally distributed in the dataset.
    \item \textbf{AG's News}\footnote{\url{http://groups.di.unipi.it/~gulli/AG_corpus_of_news_articles.html}}: The AG's News dataset consists of news articles for topic classification. The dataset has four equally distributed labels: \texttt{World}, \texttt{Sports}, \texttt{Business} and \texttt{Sci/Tech}. 
\end{itemize}

\noindent We list the details of the datasets in table~\ref{tab:dataset}. Note that in preprocessing, we limit the maximum number of words to 600 for each example in the IMDB dataset. We do not limit the maximum number of words in the AG's News dataset. Additionally, examples in both datasets are tokenized using NLTK\footnote{\url{https://www.nltk.org/}}. The average/maximum number of words is computed after preprocessing.

\noindent\textbf{Models} We consider two different DNN models to test the effectiveness of our attack. Specifically, we use word-based convolutional neural networks (CNN) and recurrent neural networks (RNN). We use LSTM as the recurrent unit in RNN. A CNN or RNN is a text encoder, which takes as input texts $\bm X$ and outputs a fixed-size vector $\bm v$. A fully connected layer with softmax activation is followed for classification. For both models, we use 100-dimensional GloVe embeddings\footnote{\url{https://nlp.stanford.edu/projects/glove/}}~\cite{glove} in our experiments. All hidden layers are 128-dimensional. 
Table~\ref{tab:clean} gives the performance of our model on clean examples. 
These results are comparable to model performance in other studies~\cite{genetic,pwws}, which means that our implementation is fair and that we are ready to investigate performance of our adversarial attacks on these models.

\begin{table}[]
\centering
\resizebox{0.3\linewidth}{!}{
\begin{tabular}{|l||l|l|}
\hline
\diagbox{\textbf{Dataset}}{\textbf{Model}}             & \textbf{CNN}   & \textbf{RNN}   \\ \hline \hline
{IMDB}       & 88.49          & 85.69 \\ \hline
{AG's News}  & 92.18          & 91.17 \\ \hline
\end{tabular}
}
\caption{Test accuracy (\%) of our model on clean examples.}
\label{tab:clean}
\end{table}

\begin{table*}[!t]
\large
\centering
\resizebox{1.0\columnwidth}{!}{
\begin{tabular}{|l||>{\centering\arraybackslash}p{2.4cm}|>{\centering\arraybackslash}p{2.2cm}|>{\centering\arraybackslash}p{2.4cm}|>{\centering\arraybackslash}p{2.2cm}||>{\centering\arraybackslash}p{2.4cm}|>{\centering\arraybackslash}p{2.2cm}|>{\centering\arraybackslash}p{2.4cm}|>{\centering\arraybackslash}p{2.2cm}|}
\hline
\multirow{3}{*}{\textbf{Method}} & \multicolumn{4}{c||}{\textbf{IMDB}}                           & \multicolumn{4}{c|}{\textbf{AG's News}}                         \\ \cline{2-5} \cline{6-9}
                       & \multicolumn{2}{c|}{\textbf{CNN}} & \multicolumn{2}{c||}{\textbf{RNN}} & \multicolumn{2}{c|}{\textbf{CNN}} & \multicolumn{2}{c|}{\textbf{RNN}} \\ \cline{2-5} \cline{6-9}
                       & \textbf{\% Replaced$^\downarrow$}  & \textbf{\% Success$^\uparrow$} & \textbf{\% Replaced$^\downarrow$}  & \textbf{\% Success$^\uparrow$}  & \textbf{\% Replaced$^\downarrow$}  & \textbf{\% Success$^\uparrow$}  & \textbf{\% Replaced$^\downarrow$}  & \textbf{\% Success$^\uparrow$}  \\ \hline \hline
\newcite{pwws}                   & 3.59       & 88.95       & 3.79       & 84.09       & \textbf{10.01}      & 85.95       & 15.33      & 79.90       \\ \hline
Our Attack           & \textbf{3.19}       & \textbf{96.12}       & \textbf{2.97}       & \textbf{99.09}       & 16.33      & \textbf{86.49}       & \textbf{14.91}      & \textbf{87.08}       \\ \hline
\end{tabular}
}

\caption{Results of adversarial attacks.  \textbf{Replaced}: Average word replacement rate. \textbf{Success}: Success rate of attack. Larger$^\uparrow$ (or lower$^\downarrow$) numbers indicate the attack is more efficient.}
\label{tab:attacks}

\end{table*}

\subsection{Adversarial Attacks}\label{sec:exp-attack}

We limit the maximum number of word replacements to 50 and 25 for the IMDB dataset and AG's News dataset, respectively. In other words, the algorithm gives up if it still cannot find an adversarial example after the number of words replaced in the original example has exceeded the limit. We report the success rate of our attack on all \textit{correctly classified} examples from the testset to prevent the model performance on clean examples from confounding the attack results. We also report the average word replacement rate for our adversarial examples. A lower word replacement rate makes it harder for humans to distinguish adversarial examples from the original ones.

We compare our attack with the state-of-the-art attack Probability Weighted Word Saliency (PWWS)~\cite{pwws}, which uses a greedy algorithm based on heuristics like word saliency and true class probability. For a fair comparison, the max length of examples is set to 600 for the IMDB dataset. We do not limit the maximum number of words for the AG's News dataset. While we obtain the results of PWWS for each dataset by evaluating on 1,000 randomly selected original examples from the testset, we evaluate our attack on the entire testset. The execution inefficiency of the original implementation from~\cite{pwws} makes it unrealistic to evaluate PWWS on the entire testset. 

Table~\ref{tab:attacks} shows the results of our adversarial attacks. As we can see, our attack outperforms PWWS in most of the metrics.
For the IMDB dataset, our attack fools the CNN and RNN model with success rates of 96.12\% and 99.09\%, respectively. Our success rates are higher than the success rates of PWWS (88.95\% for CNN, 84.09\% for RNN). While reaching higher success rates, our method also has lower word replacement rates than PWWS. The average word replacement rate is 3.19\% for CNN model and 2.97\% for RNN model, both of which are lower than results from PWWS (3.59\% for CNN, 3.79\% for RNN).

For AG's News dataset, the success rate of our method is 86.49\% for CNN and 87.08\% for RNN. The average word replacement rate is 16.33\% for CNN and 14.91\% for RNN. Our method still outperforms PWWS in the RNN model, where the success rate and average word replacement rate of PWWS are 79.90\% and 15.33\%, respectively. However, in the CNN model, while having a similar success rate, our model has a higher word replacement rate of 16.33\% compared to 10.01\% of PWWS. 

Compared to attack results of the IMDB dataset, we obtain lower success rates and higher word replacement rates for AG's News dataset. Possible explanations are: (1) fooling a multi-class classifier is harder than fooling a binary classifier, (2) examples of the IMDB dataset are longer than examples of AG's News dataset, and it is easier to generate adversarial examples for longer sequences.

Table~\ref{tab:example} gives some adversarial examples generated by our attack. As the attack replaces words from the original example, the true class probability decreases and the resulting text vector moves closer to the decision boundary. A distance smaller than 0 indicates that the text vector has crossed the decision boundary.

\begin{table*}[!t]
\centering
\resizebox{0.9\columnwidth}{!}{
\begin{tabular}{|l||>{\centering\arraybackslash}p{2cm}|>{\centering\arraybackslash}p{2cm}||>{\centering\arraybackslash}p{2cm}|>{\centering\arraybackslash}p{2cm}||>{\centering\arraybackslash}p{2cm}|>{\centering\arraybackslash}p{2cm}|}
\hline
\multirow{2}{*}{\textbf{Dataset}} & \multicolumn{2}{c||}{\textbf{\% Accuracy}} & \multicolumn{2}{c||}{\textbf{Similarity}} & \multicolumn{2}{c|}{\textbf{Modified}} \\ \cline{2-7} 
                                  & \textbf{Original}     & \textbf{Adversarial}     & \textbf{Original}     & \textbf{Adversarial}     & \textbf{Original}      & \textbf{Adversarial}     \\ \hline\hline
IMDB                              & 92                 & 90                & 4.13               & 3.40              & 2.59                & 3.14              \\ \hline
AG's News                            & 90                 & 81                & 4.96               & 3.29              & 2.18                & 3.16              \\ \hline
\end{tabular}
}

\caption{Human evaluation. \textbf{Accuracy}: prediction accuracy of human on the examples. \textbf{Similarity}: Given a score of 1-5, judge the similarity of the example text to the original text. \textbf{Modified}: Given a score of 1-5, judge the possibility of the example text having been modified by a machine. Higher score indicates higher similarity/possibility.}
\label{tab:human}
\end{table*}

\subsection{Human Evaluation}\label{sec:exp-human}
We conducted a human evaluation to understand the perceptibility of our adversarial perturbations. For each dataset, we randomly select 100 adversarial examples and the corresponding original examples. We hired workers from Amazon Mechanical Turk\footnote{\url{https://www.mturk.com/}} to conduct the evaluation. We asked the workers to perform three tasks: 

\begin{itemize}
    \item[(1)] \textbf{Accuracy}: Predict the label of an example.
    \item[(2)] \textbf{Similarity}: Judge the similarity of the given example to the original example.
    \item[(3)] \textbf{Modified}: Judge the possibility that some words in the texts having been replaced by a machine.
\end{itemize}

For the last two tasks, the workers are required to give a score between 1 to 5. A higher score indicates more similarity/higher possibility. For each task, each assignment is shown to five workers. All assignments are randomly shuffled before shown to workers. For task (1), we take the majority of the five predictions as our final label. Note that for the AG's News dataset, we count an example as incorrectly classified if no majority label exists. For tasks (2) and (3), we average scores across workers. 

Table~\ref{tab:human} shows the results of our human evaluation. For the IMDB dataset, the prediction accuracy on adversarial examples is only 2\% lower than the accuracy on original examples. This shows that adversarial examples generated by our attack mostly preserve the content and sentiment of the original examples. The evaluation on similarity and possibility of modifications also shows that the perceived difference between our adversarial examples and the original examples is relatively small. Note that the perturbations are more perceptible on AG's News dataset, which is expected as the average word replacement rate of AG's News dataset is higher than that of the IMDB dataset. 

To better understand human performance in judging similarity of texts, the workers were also asked in task (2) to give the similarity score between two identical original examples (refer to column 4 of Table~\ref{tab:human}). We see from Table~\ref{tab:human} that although the workers were expected to give a score of 5 for identical examples, they gave a score of 4.13 and 4.96 for the IMDB and AG's News dataset, respectively. The score for the IMDB dataset (4.13) is lower than that of the AG's News dataset (4.96). We believe that examples of the IMDB dataset longer than the examples of the AG's News dataset makes it harder for workers to judge whether or not two examples are identical.

\begin{figure*}[!t]
    \centering
    \resizebox{1.0\columnwidth}{!}{
        \subfloat[]{\includegraphics[width=0.34\columnwidth]{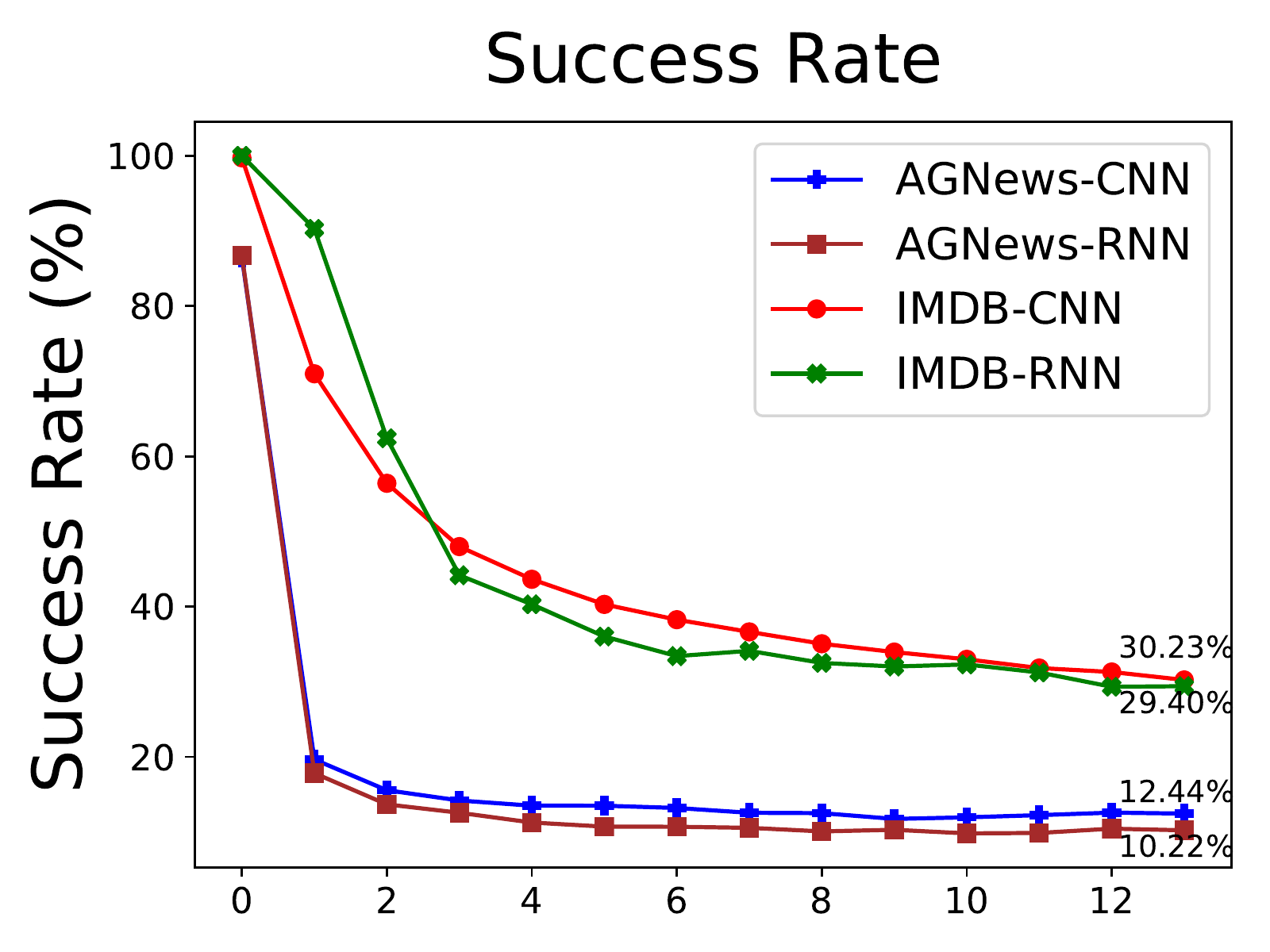}}
        \subfloat[]{\includegraphics[width=0.34\columnwidth]{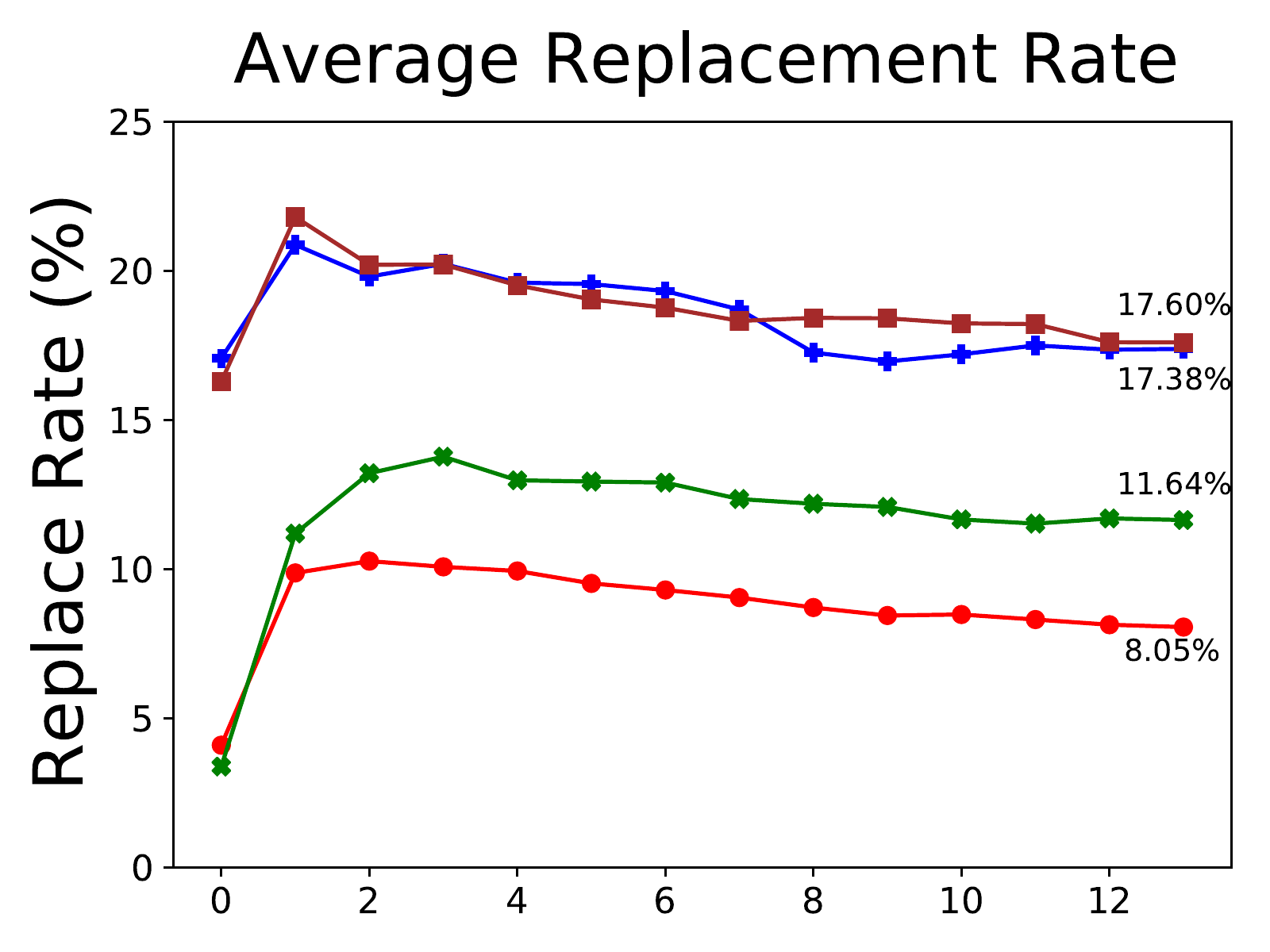}}
        \subfloat[]{\includegraphics[width=0.34\columnwidth]{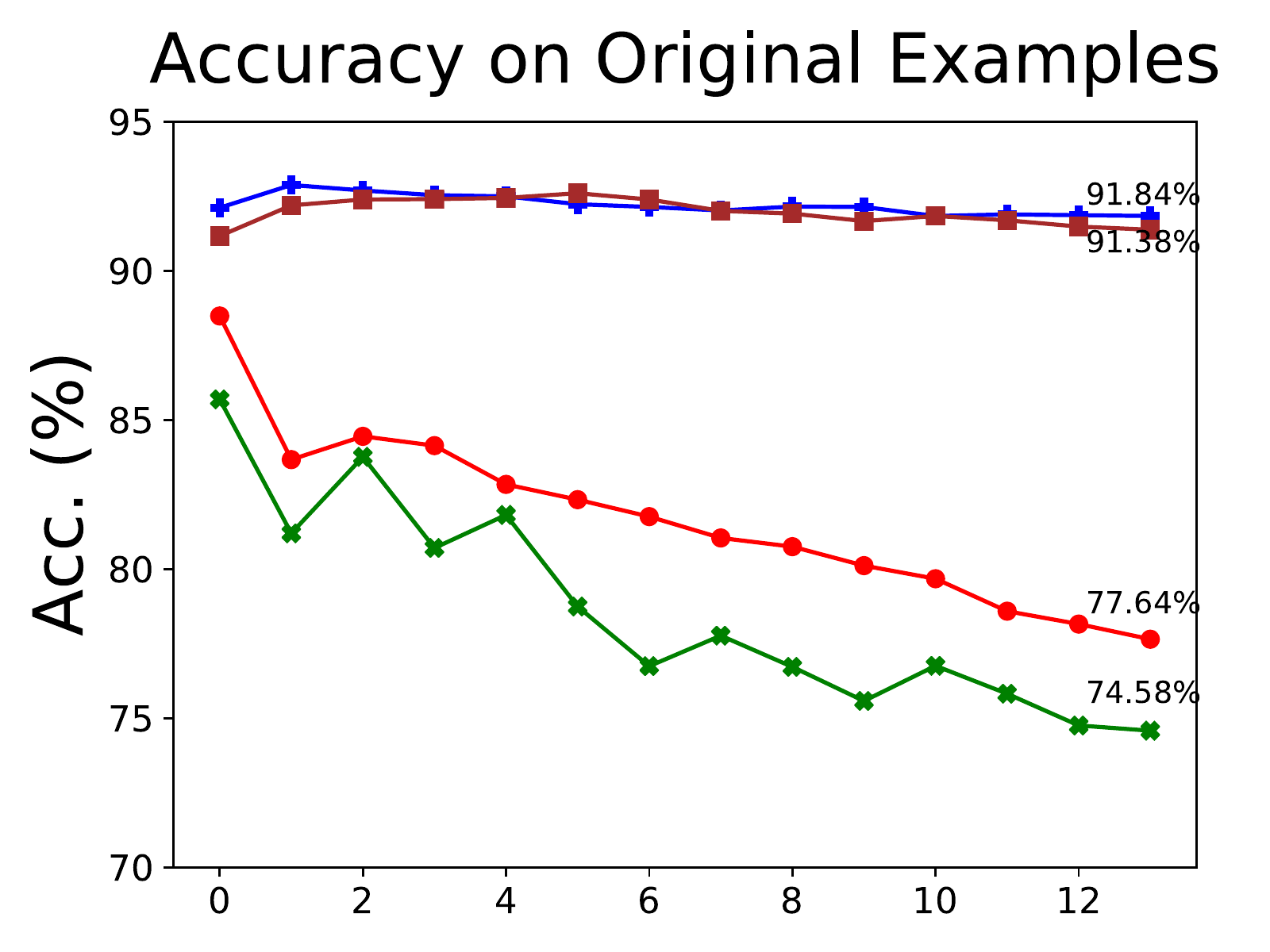}}
    }
    \caption{Results of adversarial training. The $x$-axis is epoch. Epoch 0 corresponds to models trained only on clean examples. We put $y$ values of the last epoch explicitly on the figures. (a) Success rate of our attack. (b) Average word replacement rate of adversarial examples. (c) Model performance on original examples. }
    \label{fig:adv_training}
\end{figure*}

\subsection{Adversarial Training}\label{sec:exp-training}

We conduct further experiments to validate if robustness against our attack can be achieved by adversarial training. To save time, we do adversarial training by fine-tuning on pretrained models. During adversarial training, the training set is augmented by adversarial examples, which successfully fool the model and are generated in each epoch by perturbing the \textit{correctly classified} examples. 

Figure~\ref{fig:adv_training} (a) shows that adversarial training helps, as the success rates of the attack gradually drops. Figure~\ref{fig:adv_training} (b) demonstrates that the average replacement rates increase as we do adversarial training. Lower success rates and higher word replacement rates show that the models are gaining robustness against our attack by adversarial training. We also notice from Figure~\ref{fig:adv_training} (b) that the average word replacement rates of adversarial examples start to decrease after training for some epochs. We believe that the model becomes more robust by first identifying adversarial examples with higher word replacement rates. Hence, the adversarial examples left after some epochs of adversarial training have relatively lower word replacement rates.

Figure~\ref{fig:adv_training} (c) shows the model accuracy on clean examples. For the IMDB dataset, adversarial training gradually lowers the model accuracy on clean examples. This is in line with previous image domain research, showing that model robustness is at odds with accuracy~\cite{robustness}. However, we do not observe this phenomenon for the AG's News dataset. This indicates that although adversarial training for texts and images are similar, they are still different in certain aspects.

\section{Conclusion and Future Work}~\label{sec:conclusion}
\vspace{-0.2cm}

In this paper, we propose a geometry-inspired attack for generating natural language adversarial examples. Our attack generates adversarial examples by iteratively approximating the decision boundary of Deep Neural Networks. Experiments on two text classification tasks with two models show that our attack reaches high success rates while keeping
low word replacement rates. Human evaluation shows that adversarial examples generated by our attack are hard to recognize for humans. Experiments also show that adversarial training increases model robustness against our attack.
Our current attack works for models with context-independent word embeddings. In the future, we would like to extend our attack to models using contextualized word embeddings, including ELMo~\cite{elmo}, BERT~\cite{bert}, etc.

\bibliographystyle{coling}
\bibliography{coling2020}

\end{document}